\documentclass[jmlr]{article}

\usepackage{jmlr2e}
\usepackage[colorlinks,
            linkcolor=blue,
            citecolor=blue,
            urlcolor=magenta,
            linktocpage,
            plainpages=false]{hyperref}
            
\usepackage{amsfonts}
\usepackage{amsmath,bm,paralist}
\usepackage{algorithm}
\usepackage{algorithmic}
\usepackage{epstopdf}
\usepackage{subfigure}
\usepackage[T1]{fontenc}

\makeatletter
\AtBeginDocument{\Hy@breaklinkstrue}
\makeatother

\makeatletter
\newcommand\figcaption{\def\@captype{figure}\caption}
\newcommand\tabcaption{\def\@captype{table}\caption}
\makeatother

\def \O {\mathcal{O}}

\def \R {\mathbb{R}}

\DeclareMathOperator*{\rk}{rank}
\DeclareMathOperator*{\spa}{span}
\DeclareMathOperator*{\E}{E}

\newtheorem{thm}{Theorem}
\newtheorem{lem}{Lemma}

\newtheorem{cor}{Corollary}

\title{Matrix Completion from Non-Uniformly Sampled Entries}
\usepackage{times} 
 \author{\name Yuanyu Wan \email wanyy@lamda.nju.edu.cn\\
 \addr National Key Laboratory for Novel Software Technology\\
 \addr Nanjing University, Nanjing 210023, China\\
 \name Jinfeng Yi \email yijinfeng@jd.com\\
 \addr JD AI Research, Beijing, China\\
 \name Lijun Zhang \email zhanglj@lamda.nju.edu.cn\\
 \addr National Key Laboratory for Novel Software Technology\\
 \addr Nanjing University, Nanjing 210023, China\\
 }

\begin{document}

\maketitle

\begin{abstract}
In this paper, we consider matrix completion from non-uniformly sampled entries including fully observed and partially observed columns. Specifically, we assume that a small number of columns are randomly selected and fully observed, and each remaining column is partially observed with uniform sampling. To recover the unknown matrix, we first recover its column space from the fully observed columns. Then, for each partially observed column, we recover it by finding a vector which lies in the recovered column space and consists of the observed entries. When the unknown $m\times n$ matrix is low-rank, we show that our algorithm can exactly recover it from merely $\Omega(rn\ln n)$ entries, where $r$ is the rank of the matrix. Furthermore, for a noisy low-rank matrix, our algorithm computes a low-rank approximation of the unknown matrix and enjoys an additive error bound measured by Frobenius norm. Experimental results on synthetic datasets verify our theoretical claims and demonstrate the effectiveness of our proposed algorithm.
\end{abstract}

\section{Introduction}
Recently, low-rank matrix completion has received a great deal of interests due to its theoretical advances \citep{Candes:MC:2009,MC:Keshavan}, as well as its application to a wide range of real-world problems, including recommendation \citep{Goldberg:1992:UCF,DBLP:conf/nips/YiHV0L17,yi2013inferring}, sensor networks \citep{Biswas:2006:SPB}, computer vision \citep{NIPS2011_4419}, security \citep{yi2014privacy}, human resource \citep{horesh2016information}, crowdsourcing \citep{DBLP:conf/nips/YiJJJY12,yi2012crowdclustering}, and machine learning \citep{ICML2011Jalali,yi2012robust,yi2013semi}. Let $M$ be an unknown matrix of size $m\times n$, and we assume $m \leq n$ without loss of generality. The information available about $M$ is a sampled set of entries $M_{ij}$, where $(i,j) \in \O$ and $\O\subset [m] \times [n]$.  Our goal is to recover $M$ as precisely as possible.

Most of the previous work in matrix completion assumes the entries in $\O$ are Sampled Uniformly at Random (abbr.~SUR) \citep{Candes:MC:2010,Recht:2011:SAM}. However, this assumption may be violated in real-world applications. For example, in image annotations, where the data is a matrix between images and tags, it is common to observe entire columns that correspond to well-studied categories and entire rows that correspond to labeled images. In medical diagnosis, where the data is a matrix between patients and medical measurements, it is possible to collect entire columns that correspond to inexpensive measurements and entire rows that correspond to important patients. Thus, it is natural to ask whether it is possible to recover an unknown matrix from some rows and/or columns.

Perhaps a bit surprising, answers to the above problem can be found in the recent developments of CUR matrix decomposition. The goal of CUR decomposition is to approximate a matrix $M$ by $\hat{M}=CUR$, where $U$ is estimated according to specific methods, $C$ and $R$ contain some rows and columns of $M$, respectively. While most algorithms for CUR require that $M$ is known beforehand, there are several exceptions, including a Nystr\"{o}m-type algorithm (abbr.~Nystr\"{o}m) \citep{Relative:CUR} and CUR$+$ \citep{ICML2015_Xu}.  CUR$+$ aims to recover an unknown matrix from
\begin{subequations}
\begin{align}
&\textrm{a set of columns SUR  from } [n], \label{eqn:1a} \\
&\textrm{a set of rows SUR from } [m], \textrm{and}\label{eqn:1b} \\
&\textrm{a small set of entries SUR from } [m] \times [n]. \label{eqn:1c}
\end{align}
\end{subequations}
Nystr\"{o}m can approximate the unknown matrix by only using  (\ref{eqn:1a}) and (\ref{eqn:1b}).

\begin{figure}[t]
\centering
\subfigure[Some columns are fully observed, and others are partially observed at same positions.]{\includegraphics[width=0.495\textwidth]{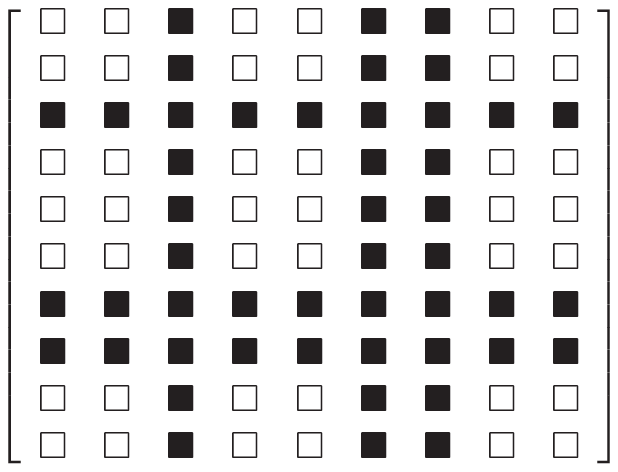}\label{fig:1:a}}
\centering
\subfigure[Some columns are fully observed, and others are partially observed at different positions.]{\includegraphics[width=0.495\textwidth]{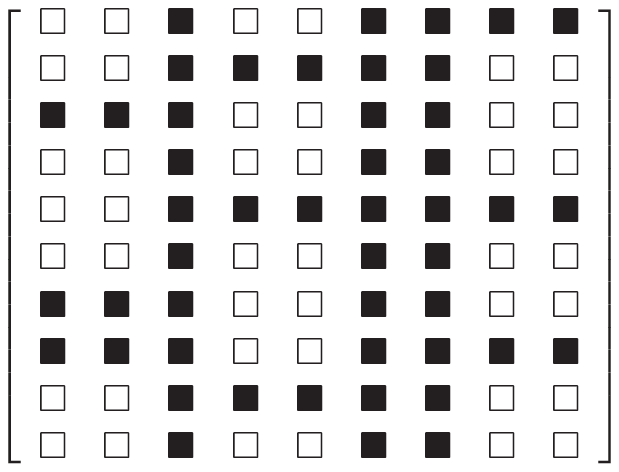}\label{fig:1:b}}
\caption{Examples of partially observed matrices. $\blacksquare$ and $\square$ indicate observed and unobserved entries, respectively.}
\label{fig:1}
\end{figure}

Along this line of research, we study the problem of matrix completion from non-uniformly sampled entries, and show that the condition in (\ref{eqn:1c}) can be dropped and the condition in (\ref{eqn:1b}) can be relaxed. Specifically, we assume that the learner observes
\begin{subequations}
\begin{align}
&\textrm{a set of columns SUR from } [n], \textrm{and}\\
& \textrm{a set of entries SUR for each remaining column}. \label{eqn:2b}
\end{align}
\end{subequations}
Note that in (\ref{eqn:2b}), we only require entries from the same column are SUR, and entries from different columns could be sampled jointly. In particular, the condition in (\ref{eqn:1b}) is a special case of (\ref{eqn:2b}). Thus, our observation model is more general than both Nystr\"{o}m \citep{Relative:CUR} and CUR$+$ \citep{ICML2015_Xu}. An illustration of our observation model is given in Figure \ref{fig:1}. While our algorithm can handle both cases in Figure \ref{fig:1:a} and Figure \ref{fig:1:b}, Nystr\"{o}m is limited to Figure \ref{fig:1:a}, and CUR$+$ requires even more information than Nystr\"{o}m.

Our recovery algorithm consists of two simple steps:
\begin{compactenum}
\item recover the column space of $M$ from the observed columns, and
\item recover each partially observed column by finding a vector that lies in the recovered column space and consists with the observed entries.
\end{compactenum}
Let $r$ be the rank of the unknown matrix $M$. When $M$ is of low-rank and satisfies the incoherence condition, it can be recovered perfectly when we observe $\Omega(r \ln r)$ columns completely and $\Omega(r \ln n)$ entries for each remaining column.  Thus, the sample complexity is $\Omega(r n \ln n)$, which is slightly better than the $\Omega( r n \ln^2 n)$ of the conventional matrix completion \citep{Recht:2011:SAM}. Furthermore, when the unknown matrix $M$ is the sum of a low-rank matrix $C$ and a Gaussian noise matrix $R$, we establish an upper bound measured by Frobenius norm for recovering its best rank-$r$ approximation $M_r$, under the assumption that the top eigenspaces of $C$ and the column norm of $M$ are incoherent.

\section{Related Work}
In this section, we briefly review the related work in matrix completion and low-rank matrix approximation.

\noindent{\textbf{Matrix Completion}\ \ \
Matrix completion refers to the problem of recovering a low-rank matrix based on partially observed entries, and motivates a series of work \citep{Candes:MC:2009,Candes:MC:2010,MC:Keshavan,Gross2011,Recht:2011:SAM} which can exactly recover a rank-$r$ matrix of size $m\times n$ from $\Omega( r n \ln^2 n)$ uniformly observed entries based on the framework of convex optimization under the incoherence condition. Recently, two improvements \citep{NIPS2013_Krishnamurthy,ICML2015_Xu} have been proposed to further reduce the sample complexity by adopting better strategies to observe necessary entries rather than uniform sampling from the whole unknown matrix.

Specifically, the algorithm proposed by \citet{NIPS2013_Krishnamurthy} needs to observe a set of columns sampled with adaptive sensing strategy and a set of rows sampled uniformly at random, and its sample complexity is $\Omega(r^{3/2}n\ln r)$. We note that two recent work \citep{Akshay2014,Balcan16} improves the sample complexity of this algorithm to $\Omega(rn\ln^2 r)$ and $\Omega(rn\ln r)$ respectively. Although its sample complexity may be slightly better than our $\Omega(r n \ln n)$, this algorithm cannot handle the observation model in Figure \ref{fig:1:b}. As mentioned before, CUR$+$ \citep{ICML2015_Xu} requires more conditions as it needs to observe a set of entries, columns, and rows sampled uniformly at random. Compared with our algorithm, the observation model of CUR+ is too restrictive, although it has a slightly better sample complexity.

\noindent{\textbf{Low-Rank Matrix Approximation}\ \ \
Low-rank matrix approximation refers to the problem of approximating a given $m\times n$ matrix by another matrix of rank not greater than a specified rank $r$ where $r\ll\min(m,n)$. It arises from numerous applications such as latent semantic indexing \citep{latent90}, DNA microarray analysis \citep{DNA01}, face recognition \citep{face91}, and web search  \citep{web99}. Although singular value decomposition (SVD) can be used to find an optimal solution for this problem in a well-defined sense, it is not scalable since its memory and time complexities are superlinear w.r.t. $m$ and $n$. To address this issue, several efficient algorithms based on norm sampling \citep{LRA04,Drineas_2006_II} and adaptive sampling \citep{SODA06,LRA06} have been proposed. Although these algorithms can find a good low-rank approximation, they need to observe the entire matrix.

To deal with missing data, \citet{LRA07} propose entrywise subsampling whose main idea is to construct an unbiased estimator of the unknown matrix and compute the SVD of the estimator. This algorithm has shortcomings that it cannot exactly recover a low-rank matrix and cannot capture the real column space of the unknown matrix. Although Nystr\"{o}m \citep{Relative:CUR} can capture the actual column and row space of the unknown matrix, it is limited to solve the problem illustrated in Figure \ref{fig:1:a}. CUR$+$, which requires more restrictive observation model, focus on the setting that the unknown matrix has skewed singular value distribution when it is of full-rank. Even so, CUR$+$ requires nearly the entire matrix for finding a good low-rank approximation, as discussed by \citet{ICML2015_Xu}. The adaptive strategy and its variant have also been applied to computing a better low-rank approximation for an unknown noisy low-rank matrix \citep{NIPS2013_Krishnamurthy,Wang2015}. However, these algorithms rely on sampling strategies that require more than one pass over the unknown matrix and their bounds have a poor dependence on failure probability $\delta$, i.e., $\Omega(1/\delta)$, which significantly limits their applications when columns are uniformly sampled, or the unknown matrix can only be sampled with one pass.

\section{Main Results}
In this section, we present our algorithms and the corresponding theoretical results.
\subsection{The Proposed Algorithm}
For a matrix $B\in\mathbb{R}^{m\times n}$, let $B_{(i)}$ and $B^{(j)}$ denote the $i$-th row and $j$-th
column of $B$, respectively. For a set $\O \subset[m]$, the vector $\mathbf{x}_\O\in \mathbb{R}^{|\O|}$ contains elements of vector $\mathbf{x}$ indexed by $\O$. Similarly the matrix {$B_\O\in \mathbb{R}^{|\O|\times n}$} has rows of matrix $B$ indexed by $\O$.

Let $M=[\mathbf{m}_1,\mathbf{m}_2,\cdots,\mathbf{m}_n]\in\mathbb{R}^{m\times n}$ be the matrix to be recovered.  Let $\{p_i\}_{i=1}^n$ be a probability distribution used to randomly sample columns where $p_i>0,\ \sum_{i=1}^np_i=1$. To approximate $M$, we first sample $d$ columns from $M$ and construct $A\in \mathbb{R}^{m\times d}$, where we pick $i_j\in\left\{1,...,n\right\}$ with $\Pr\left[i_j=i\right]=p_i$ and set $A^{(j)}=M^{(i_j)}/\sqrt{dp_{i_j}}$ for $j=1,\cdots,d.$
Let $r\leq d$ be the target rank and $\hat{r}=\min\left(r,\rk(A)\right)$. We calculate the top-$\hat{r}$ left singular vectors of $A$ denoted by 
$\hat{U}=[\hat{\mathbf{u}}_1,\hat{\mathbf{u}}_2,\cdots,\hat{\mathbf{u}}_{\hat{r}}]$ that is approximately the column space of $M$. For each of the rest columns $\mathbf{m}_i$, we sample a set $\O_{i}$ of $s$ entries uniformly at random with replacement, denoted by $\mathbf{m}_{i,\O_{i}}$. We then solve the following optimization problem
\begin{equation}
\label{opt}
\min_{\mathbf{z}\in\R^r}\frac{1}{2}\|\mathbf{m}_{i,\O_{i}}-\hat{U}_{\O_{i}}\mathbf{z}\|_2^2
\end{equation}
to recover this column by $\hat{\mathbf{m}}_i=\hat{U}\mathbf{z}_\ast$, where $\mathbf{z}_\ast$ is the optimal solution. Because the problem (\ref{opt}) has a  closed-form solution $\mathbf{z}_\ast=(\hat{U}_{\O_{i}}^T\hat{U}_{\O_{i}})^{-1}\hat{U}_{\O_{i}}^T\mathbf{m}_{i,\O_{i}}$, we have \[\hat{\mathbf{m}}_i=\hat{U}(\hat{U}_{\O_{i}}^T\hat{U}_{\O_{i}})^{-1}\hat{U}_{\O_{i}}^T\mathbf{m}_{i,\O_{i}}.\]

The detailed procedures are summarized in Algorithm \ref{LR}.
\begin{algorithm}[t]
\caption{Matrix Completion from Non-Uniformly Sampled Entries}
\label{LR}
\begin{algorithmic}[1]
\STATE \textbf{Input:} $r>0$, $d>0$, $s>0$, $\{p_i\}_{i=1}^n$ where $p_i>0,\sum_{i=1}^np_i=1$
\FOR{$j=1,\cdots,d$}
\STATE Sample $i_j\in[n]$ with $\Pr\left[i_j=i\right]=p_i$
\STATE Set $\hat{\mathbf{m}}_{i_j}=M^{(i_j)}$ and $A^{(j)}=M^{(i_j)}/\sqrt{dp_i}$
\ENDFOR
\STATE Set $\hat{r}=\min\left(r,\rk(A)\right)$
\STATE Calculate the top-$\hat{r}$ left singular vectors of $A$ denoted by $\hat{U}=[U^{(1)},U^{(2)},\cdots,U^{(\hat{r})}]$
\FOR{each of the rest columns $\mathbf{m}_i$}
\STATE Sample a set $\O_{i}$ of $s$ entries uniformly at random with replacement denoted by $\mathbf{m}_{i,\O_{i}}$
\STATE Calculate $\hat{\mathbf{m}}_i =
\hat{U}(\hat{U}_{\O_i}^T\hat{U}_{\O_i})^{-1}\hat{U}_{\O_i}^T\mathbf{m}_{i,\O_i}$
\ENDFOR
\STATE \textbf{Output: $\hat{M} = [\hat{\mathbf{m}}_1,\cdots,\hat{\mathbf{m}}_n]$}
\end{algorithmic}
\end{algorithm}

\subsection{Theoretical Guarantees}
Given $r\in[n]$, let $\bar{U} = [\mathbf{u}_1,\mathbf{u}_2,\cdots,\mathbf{u}_r]\in\mathbb{R}^{m\times r}$ and $\bar{V} = [\mathbf{v}_1,\mathbf{v}_2,\cdots,\mathbf{v}_r]\in\mathbb{R}^{n\times r},$ where $\{\mathbf{u}_i\}_1^{r}$ and $\{\mathbf{v}_i\}_1^{r}$ are the top-$r$ left and right singular vectors of $M$, respectively.
Define projection operators $P_{\bar{U}}=\bar{U}\bar{U}^T$, $P_{\hat{U}}=\hat{U}\hat{U}^T$.
The incoherence measure for $\bar{U}$ and $\bar{V}$ is defined as
\begin{equation*}
\mu(r)=\max\left(\max\limits_{i\in[m]}\frac{m}{r}\|\bar{U}_{(i)}\|^2_2,\max\limits_{i\in[n]}\frac{n}{r}\|\bar{V}_{(i)}\|_2^2\right).
\end{equation*}
Similarly, the incoherence measure for $\hat{U}$ is defined as
\begin{equation*}
\hat{\mu}(\hat{r})=\max\limits_{i\in[m]}\frac{m}{\hat{r}}\|\hat{U}_{(i)}\|_2^2.
\end{equation*}
In the following, we first consider the low-rank case where $\rk(M)= r$, and then prove a general result for any fixed probability distribution $\{p_i\}_{i=1}^n$.
\begin{thm}
\label{thm1}
Let $\{p_i\}_{i=1}^n$ be the probability distribution used to randomly sample columns and $p_{\min}=\min\limits_{i\in[n]}p_i$, where $p_i>0,\ \sum_{i=1}^np_i=1$. Assume $\rk(M)= r$,\ $d\geq7\mu(r)r\ln(2r/\delta)/(np_{\min})$, and $s\geq7\mu(r)r\ln(2rn/\delta)$. With a probability at least $1-\delta$, Algorithm \ref{LR} recovers $M$ exactly.
\end{thm}
From Theorem \ref{thm1}, we find that the lower bound of $d$ depends on the probability distribution. To minimize the threshold $7\mu(r)r\ln(2r/\delta)/(np_{\min})$, we set \begin{equation}
\label{best_p}
p_{\min}=p_1=p_2=\cdots=p_n=1/n
\end{equation}
which corresponds to uniform sampling. From this perspective, uniform sampling is a useful strategy for recovering the low-rank matrix though it is very simple. By combining (\ref{best_p}) and Theorem \ref{thm1}, we provide the following corollary for our Algorithm \ref{LR}.
\begin{cor}
\label{cor1}
Assume $\rk(M)= r$, $d\geq7\mu(r)r\ln(2r/\delta)$ and $s\geq7\mu(r)r\ln(2rn/\delta)$. With a probability at least $1-\delta$, Algorithm \ref{LR} with uniform sampling recovers $M$ exactly.
\end{cor}

Corollary 1 implies the sample complexity of Algorithm \ref{LR} with uniform sampling is $\Omega(r n \ln n)$. Although it is slightly worse than the previous best result $\Omega(r n \ln r)$ \citep{ICML2015_Xu,Balcan16}, it is much more general. Besides, it is easy to verify that our sample complexity can be further reduced to $\Omega(r n \ln r)$ too when we sample the same $O_i$ as \citet{ICML2015_Xu} or \citet{Balcan16}.

In practice, the low-rank matrix may be corrupted by noise, and the matrix could be of full-rank. To handle the general setting, we assume that $M=C+R$ where $C\in\R^{m\times n}$ is a low-rank matrix and $R\in\R^{m\times n}$ is a random matrix with entries independently drawn from $\mathcal{N}(0,\sigma^2)$. Let $\bar{U} = [\mathbf{u}_1,\mathbf{u}_2,\cdots,\mathbf{u}_r]\in\mathbb{R}^{m\times r}$ where $\{\mathbf{u}_i\}_1^{r}$ are the top-$r$ left singular vectors of $C$. Our goal is to calculate a low-rank approximation of $M$. To this end, we introduce a new incoherence measure $\mu(M)=\frac{n\max_{i\in [n]}\|\mathbf{m}_i\|_2^2}{\|M\|_F^2}.$
The following theorem establishes the error guarantee of Algorithm \ref{LR}.
\begin{thm}
\label{thm2}
Assume that $M=C+R$ where $\rk(C)=r$ and $R\in\R^{m\times n}$ is a random matrix with entries independently drawn from $\mathcal{N}(0,\sigma^2)$.  Suppose that $\ln(2n/\delta)\leq m/64$ and $r\leq m/4$. Let $\hat{M}$ be the output of our Algorithm \ref{LR} with uniform sampling. Then with probability at least $ 1- \delta$, we have
\[\|M-\hat{M}\|^2_F\leq\|M-M_r\|_F^2+\epsilon\|M\|_F^2\]
provided that $d=\Omega\left(r\mu(M)\ln(1/\delta)/\epsilon^2\right)$ and $s=\Omega\left(r^2\mu(r)\ln^2(2rn/\delta)/\epsilon\right)$.
\end{thm}
Theorem \ref{thm2} shows that with an $\Omega(r^2n\mu(M)\mu(r)\ln^2(rn)/\epsilon^2)$ observation, Algorithm \ref{LR} with uniform sampling can achieve $\epsilon$ additive approximation error with an overwhelming probability. 
For comparison, we note that \citet{Wang2015} achieve a relative error guarantee as
\[\|M-\hat{M}\|^2_F\leq\frac{2.5^r(r+1)!}{\delta}\|M-M_r\|_F^2\]
with a probability $ 1- \delta$ and sample complexity $\Omega\left(r^2n\mu(r)\ln^2(n/\delta)\right)$.
However, their bounds have a poor dependence on the failure probability $\delta$, i.e., $\Omega(1/\delta)$, and their algorithm requires more than one pass over the unknown matrix. \citet{ICML2015_Xu} show that CUR$+$  achieves a similar relative error guarantee measured by spectral norm with high probability. Nevertheless, even the unknown matrix has skewed singular value distribution, CUR$+$ requires nearly the entire matrix, i.e., $\Omega(n^2/d^2)$ observed entries.

\section{Analysis}
In this section, we prove Theorems \ref{thm1} and \ref{thm2} by introducing several key lemmas. We defer the detailed proofs to the supplementary material due to space limitation.
\subsection{Proof of Theorem \ref{thm1}}
For each column $\mathbf{m}_i$ not included in $A$, if $\mathbf{m}_i\in\hat{U}$ and $\hat{U}_{\O_i}^T\hat{U}_{\O_i}$ is
invertible, we can write $\mathbf{m}_i=\hat{U}\mathbf{b}_i$, where $\mathbf{b}_i\in\R^{\hat{r}\times1}$. Thus, we have
\begin{eqnarray*}
\hat{\mathbf{m}}_i =
\hat{U}(\hat{U}_{\O_i}^T\hat{U}_{\O_i})^{-1}\hat{U}_{\O_i}^T\hat{U}_{\O_i}\mathbf{b}_i=\hat{U}\mathbf{b}_i=\mathbf{m}_i.
\end{eqnarray*}
This means that each column $\mathbf{m}_i$ can be recovered exactly under two conditions: $\mathbf{m}_i\in\hat{U}$ and $\hat{U}_{\O_i}^T\hat{U}_{\O_i}$ is
invertible. Therefore, based on the following two lemmas, we show that our Algorithm \ref{LR} with the assumptions in Theorem \ref{thm1} satisfies these two conditions.
\begin{lem}
\label{thm2-3}
Let $\{p_i\}_{i=1}^n$ be the probability distribution used to randomly sample column and $p_{\min}=\min\limits_{ i\in [n]}p_i$, where $p_i>0,\ \sum_{i=1}^np_i=1$. Assume $M$ has rank $r$, with a probability at least $1-e^{-t}$, we have
\begin{eqnarray*}
\rk(A) = r
\end{eqnarray*}
provided that $d\geq7\mu(r)r(t+\ln r)/(np_{\min}).$
\end{lem}

\begin{lem}
\label{thm2-4}
With a probability at least $1-\frac{1}{n}e^{-t}$, we have
\begin{eqnarray*}
\lambda_{\min}\left(\hat{U}_{\O_i}^T\hat{U}_{\O_i}\right)\geq\frac{|\O_i|}{2m}
\end{eqnarray*}
provided that $|\O_i|\geq7\hat{\mu}(\hat{r})\hat{r}(t+\ln \hat{r}+\ln n)$.
\end{lem}
By combining Lemma \ref{thm2-3} and the fact that $M$ has rank $r$, when $d\geq7\mu(r)r(t+\ln r)/(np_{\min})$ and with a probability at least $1-e^{-t}$, we have $\rk(A) = \rk(M)=r$, which means $\hat{r}=\min\left(r,\rk(A)\right)=r$. Note that $A$ is composed of $d$
selected and rescaled columns of $M$. Hence {$P_{\hat{U}}=P_{\bar{U}}$}, which directly implies that
${\hat{\mu}(\hat{r})}=\mu(r)$ and $\mathbf{m}_i\in\hat{U}$, $i\in[n]$. Then, according to Lemma \ref{thm2-4}
and the union bound, we have
$\lambda_{\min}\left(\hat{U}_{\O_i}^T\hat{U}_{\O_i}\right)\geq\frac{|\O_i|}{2m}$
with probability at least $ 1-e^{-t}$ for all column $\mathbf{m}_i$ with the fact
$|\O_i|=s\geq7\mu(r)r(t+\ln r+\ln n)$. Note that this means all
$\hat{U}_{\O_i}^T\hat{U}_{\O_i}$ are invertible with probability at least $ 1-e^{-t}$. Using union
bound again, we can exactly recover $M$ with a probability at least $1-2e^{-t}$. Let $\delta=2e^{-t}$, we get $t=\ln(2/\delta)$.

\subsection{Proof of Theorem \ref{thm2}}
Let $\mathbf{m}_i=\mathbf{c}+\mathbf{r}$, where $\mathbf{c}=P_{\hat{U}}\mathbf{m}_i$ and $\mathbf{r}=P_{\hat{U}^\bot}\mathbf{m}_i$. Besides Theorem \ref{thm2-4}, we further introduce several lemmas that are central to our analysis.
\begin{lem}
(Lemma $17$ in \citet{NIPS2013_Krishnamurthy})
\label{wang}
For a vector $\mathbf{x}$, let
$\mu(\mathbf{x}) = \frac{m\|\mathbf{x}\|^2_{\infty}}{\|\mathbf{x}\|_2^2}.$ Let $\mathbf{m}_i=\mathbf{c}+\mathbf{r}$, where $\mathbf{c}=P_{\hat{U}}\mathbf{m}_i$ and $\mathbf{r}=P_{\hat{U}^\bot}\mathbf{m}_i$. With probability at least $ 1-\delta$, we have
\[\|\hat{U}_\O^T\mathbf{r}_\O\|_2^2\leq\beta\frac{|\O|}{m}\frac{\hat{r}\hat{\mu}(\hat{r})}{m}\|\mathbf{r}\|_2^2\]
where $\beta = 6\ln(\hat{r}/\delta)+\frac{4}{3}\frac{\hat{r}\mu(\mathbf{r})}{|\O|}\ln^2(\hat{r}/\delta)$.
\end{lem}
\begin{lem}
\label{n1}
For a vector $\mathbf{x}$, let
$\mu(\mathbf{x}) = \frac{m\|\mathbf{x}\|^2_{\infty}}{\|\mathbf{x}\|_2^2}.$
Assume that $M=C+R$ where $\rk(C)=r$ and $R\in\R^{m\times n}$ is a random matrix with entries independently drawn from $\mathcal{N}(0,\sigma^2)$. Suppose $r\leq d\leq m/4$,\ $\ln(2n/\delta)\leq m/64$. With probability at least $1-\delta$, we have
\begin{equation}
\label{mu1}
\hat{\mu}(\hat{r})=O\left(\frac{r\mu(r)\ln(m/\delta)}{\hat{r}}\right)
\end{equation}
and with probability at least $1-\delta$,  we have
\begin{equation}
\label{mu3}
\mu(P_{\hat{U}^\bot}\mathbf{m}_i)=O\left(r\mu(r)+\ln(mn/\delta)\right)
\end{equation}
for all partially observed $\mathbf{m}_i$.
\end{lem}

\begin{lem}
\label{lemLR}
When Algorithm \ref{LR} adopts uniform sampling, with a probability at least $ 1-\delta$, we have
\[\|M-\hat{U}\hat{U}^TM\|_F^2\leq\|M-M_r\|_F^2+\epsilon\|M\|_F^2\]
provided $d\geq16\ln(2/\delta)\mu(M)r/\epsilon^2$.
\end{lem}
Due to Lemma \ref{thm2-4} and the union bound, with a probability at least $ 1-e^{-t}$, we have
\begin{eqnarray*}
\lambda_{\min}\left(\hat{U}_{\O_i}^T\hat{U}_{\O_i}\right)\geq\frac{|\O_i|}{2m}
\end{eqnarray*}
provided that $|\O_i|\geq7\hat{\mu}(\hat{r})\hat{r}(t+\ln \hat{r}+\ln n)$ for all $i\in[n]$.
So with a probability at least $1-\delta$, for all columns, we have
\begin{align*}
\|\mathbf{m}-\hat{U}(\hat{U}_{\O_{i}}^T\hat{U}_{\O_{i}})^{-1}\hat{U}_{\O_{i}}^T\mathbf{m}_{i,\O_{i}}\|_2^2=&\|\mathbf{c}+\mathbf{r}-\hat{U}(\hat{U}_{\O_{i}}^T\hat{U}_{\O_{i}})^{-1}\hat{U}_{\O_{i}}^T(\mathbf{c}_{\O_{i}}+\mathbf{r}_{\O_{i}})\|_2^2\\
=&\|\mathbf{c}+\mathbf{r}-\hat{U}(\hat{U}_{\O_{i}}^T\hat{U}_{\O_{i}})^{-1}\hat{U}_{\O_{i}}^T(\hat{U}_{\O_{i}}\hat{U}^T\mathbf{m}_{i}+\mathbf{r}_{\O_{i}})\|_2^2\\
=&\|\mathbf{r}-\hat{U}(\hat{U}_{\O_{i}}^T\hat{U}_{\O_{i}})^{-1}\hat{U}_{\O_{i}}^T\mathbf{r}_{\O_{i}}\|_2^2\\
\leq&\|\mathbf{r}\|^2_2+\|(\hat{U}_{\O_{i}}^T\hat{U}_{\O_{i}})^{-1}\|^2\|\hat{U}_{\O_{i}}^T\mathbf{r}_{\O_{i}}\|_2^2\\
\leq&\left(1+\beta\frac{4m^2}{|\O_i|^2}\frac{|\O_i|}{m}\frac{\hat{r}\hat{\mu}(\hat{r})}{m}\right)\|\mathbf{r}\|_2^2\leq\left(1+\frac{\epsilon}{3}\right)\|\mathbf{r}\|_2^2
\end{align*}
provided that $|\O_i|\geq\max\left(84\hat{\mu}(\hat{r})\hat{r}\ln(2\hat{r}n/\delta)/\epsilon,\frac{4}{3}\hat{r}\mu(\mathbf{r})\ln(2\hat{r}n/\delta)\right).$

Summing over all columns, we have
\[\|M-\hat{M}\|^2_F\leq\left(1+\frac{\epsilon}{3}\right)\|M-P_{\hat{U}}M\|_F^2.\]
According to Lemma \ref{lemLR}, if $d\geq64\ln(2/\delta)\mu(M)r/\epsilon^2$, then with probability at least $1-\delta$,
\[\|M-\hat{U}\hat{U}^TM\|_F^2\leq\|M-M_r\|_F^2+\frac{\epsilon}{2}\|M\|_F^2,\]
which leads to
\[\|M-\hat{M}\|^2_F\leq\|M-M_r\|_F^2+\epsilon\|M\|_F^2.\]
Now we check whether the conditions of $d$ and $s$ are satisfied.
First, the condition $d\geq64\ln(2/\delta)\mu(M)r/\epsilon^2$ means $d=\Omega\left(r\mu(M)\ln(1/\delta)/\epsilon^2\right)$. Then, the condition $s\geq\max\left(84\hat{\mu}(\hat{r})\hat{r}\ln(2\hat{r}n/\delta)/\epsilon,\frac{4}{3}\hat{r}\mu(\mathbf{r})\ln(2\hat{r}n/\delta)\right)$ can further derive $s=\Omega\left(r^2\mu(r)\ln^2(2rn/\delta)/\epsilon\right)$ since
\begin{eqnarray}
\hat{\mu}(\hat{r})\hat{r}\ln(2\hat{r}n/\delta)\!\!&\leq&\!\!O\left(r\mu(r)\ln^2(2\hat{r}n/\delta)\right),\nonumber\\
\hat{r}\mu(\mathbf{r})\ln(2\hat{r}n/\delta)=O\left(\hat{r}(r\mu(r)+\ln(mn/\delta))\ln(2\hat{r}n/\delta)\right)\!\!&\leq&\!\!O\left(r^2\mu(r)\ln^2(2rn/\delta)\right).\nonumber
\end{eqnarray}
\section{Experiments}
In this section, we first verify the theoretical result in Corollary \ref{cor1}, i.e., Algorithm \ref{LR} with uniform sampling has a dependence of
sample complexity on $r$ and $n$. 
To this end, we evaluate the performance of Algorithm \ref{LR} with uniform sampling by comparing it against Nystr\"{o}m \citep{Relative:CUR} for computing a low-rank approximation to a noisy low-rank matrix.
\subsection{Verifying the Dependences on $r$ and $n$}
We will verify the sample complexity in Corollary \ref{cor1}, i.e., $d\geq
\Omega(r\ln r)$ and $s\geq \Omega(r\ln n)$.
\paragraph{Settings} Here we adopt the similar settings as in \citet{ICML2015_Xu}. We study square matrices
of different sizes and ranks, with $n$ varied in $\{2000, 4000, 6000, 8000, 10000\}$, and $r$ varied in
$\{10, 20, 30, 40, 50\}$. For each special $n$ and $r$, we search for the smallest $q$ and $s$ that
can lead to almost exact recovery of the target matrix, i.e., $\|M-\hat{M}\|_F/\|M\|_F\leq 10^{-8}$ in
all $10$ independent trials. To create the rank-$r$ matrix $M\in\mathbb{R}^{n\times n}$, we randomly
generate matrix $M_L\in\mathbb{R}^{n\times r}$ and $M_R\in\mathbb{R}^{r\times n}$, where each entry of
$M_L$ and $M_R$ is drawn independently at random from $\mathcal{N}(0,1)$, and $M$ is given by
$M=M_L\times M_R$. Under this construction scheme, the difference among the incoherence $\mu(r)$ for
different sized matrices is relatively small (from $1.9049$ to $4.1616$ ), so we
ignore the impact of $\mu(r)$ in our analysis.

\paragraph{Results} The dependence of minimal $d$ on $r$ and $n$ is shown in Figures \ref{fig1}(a) and (b),
which plot $d$ against $r\ln r$ and $r^2\ln r$, respectively. We can find that $d$ has a linear
dependence on $r\ln r$ instead of $r^2\ln r$. In addition, we also find that $d$ is almost independent
from the matrix size $n$. Figures \ref{fig1}(c) and (d) plot $s$, the minimum number of
observed entries for each column, against $r\ln r$ and $r^2\ln r$. We can see that
$s$ also has a linear dependence on $r\ln r$ instead of $r^2\ln r$. According to Theorem \ref{thm1}, the requirement on $s$ depends on $\ln n$, but we find that $s$ is almost independent
from the matrix size $n$. This suggests that $s\geq \Omega(r\ln n)$ is the worst sample complexity, where $\ln n$ is caused by the union bound, and $s$ could be independent from $n$ in practice.
\begin{figure}[t]
\centering
\subfigure[$d$ against $r\ln r$]{\includegraphics[width=0.45\textwidth]{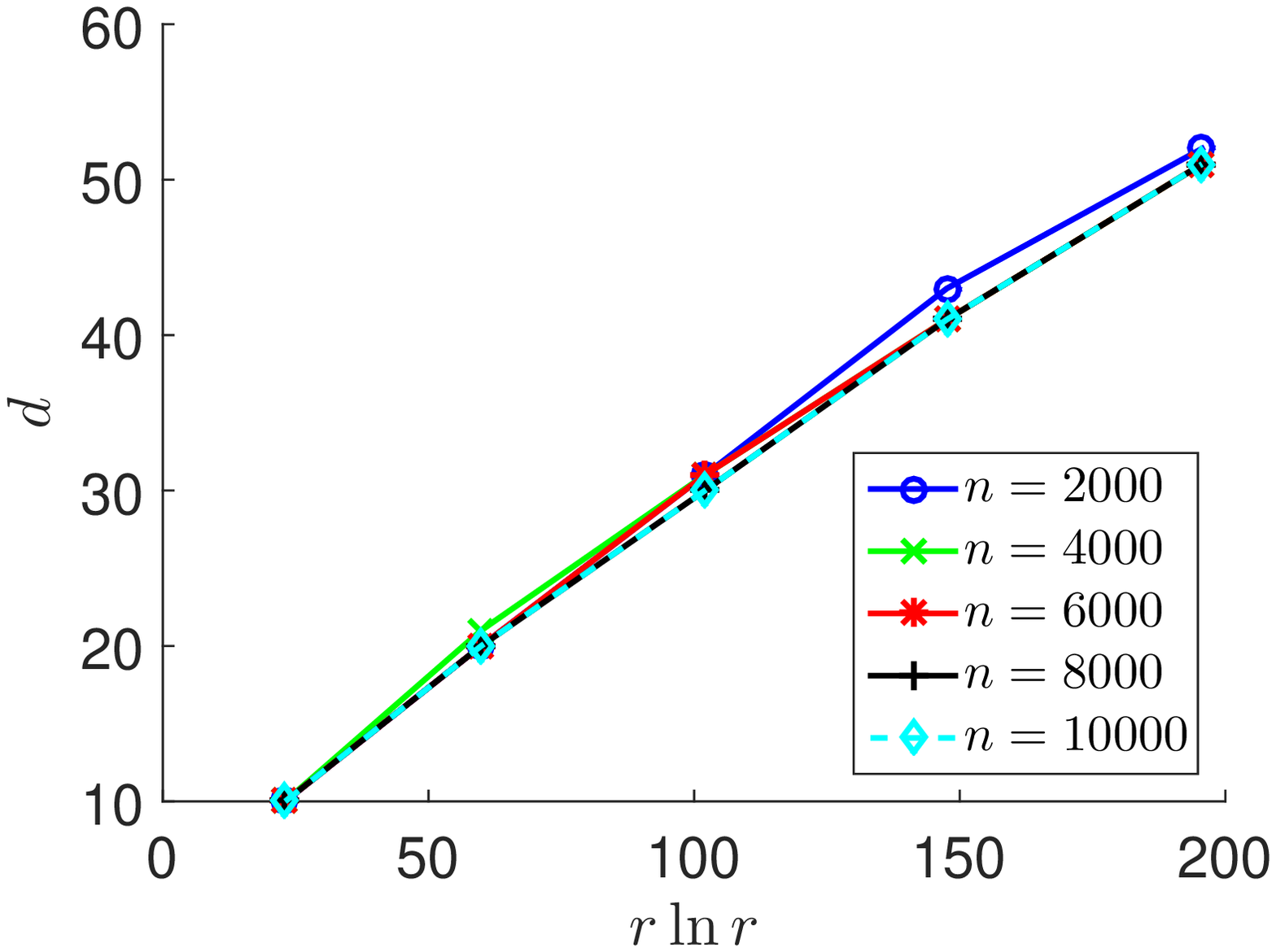}}
\centering
\subfigure[$d$ against $r^2\ln r$]{\includegraphics[width=0.45\textwidth]{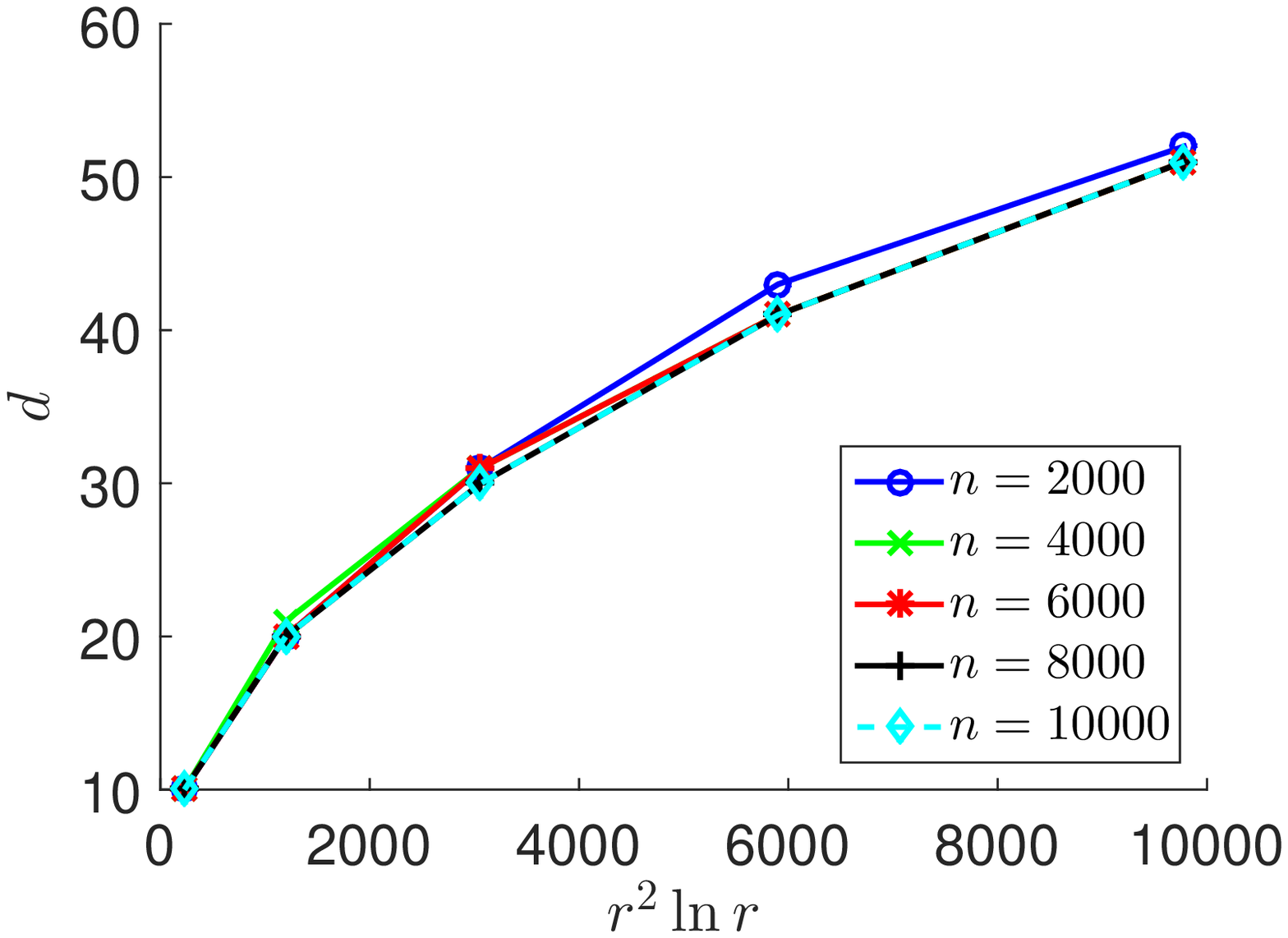}}\\
\centering
\subfigure[$s$ against $r\ln r$]{\includegraphics[width=0.45\textwidth]{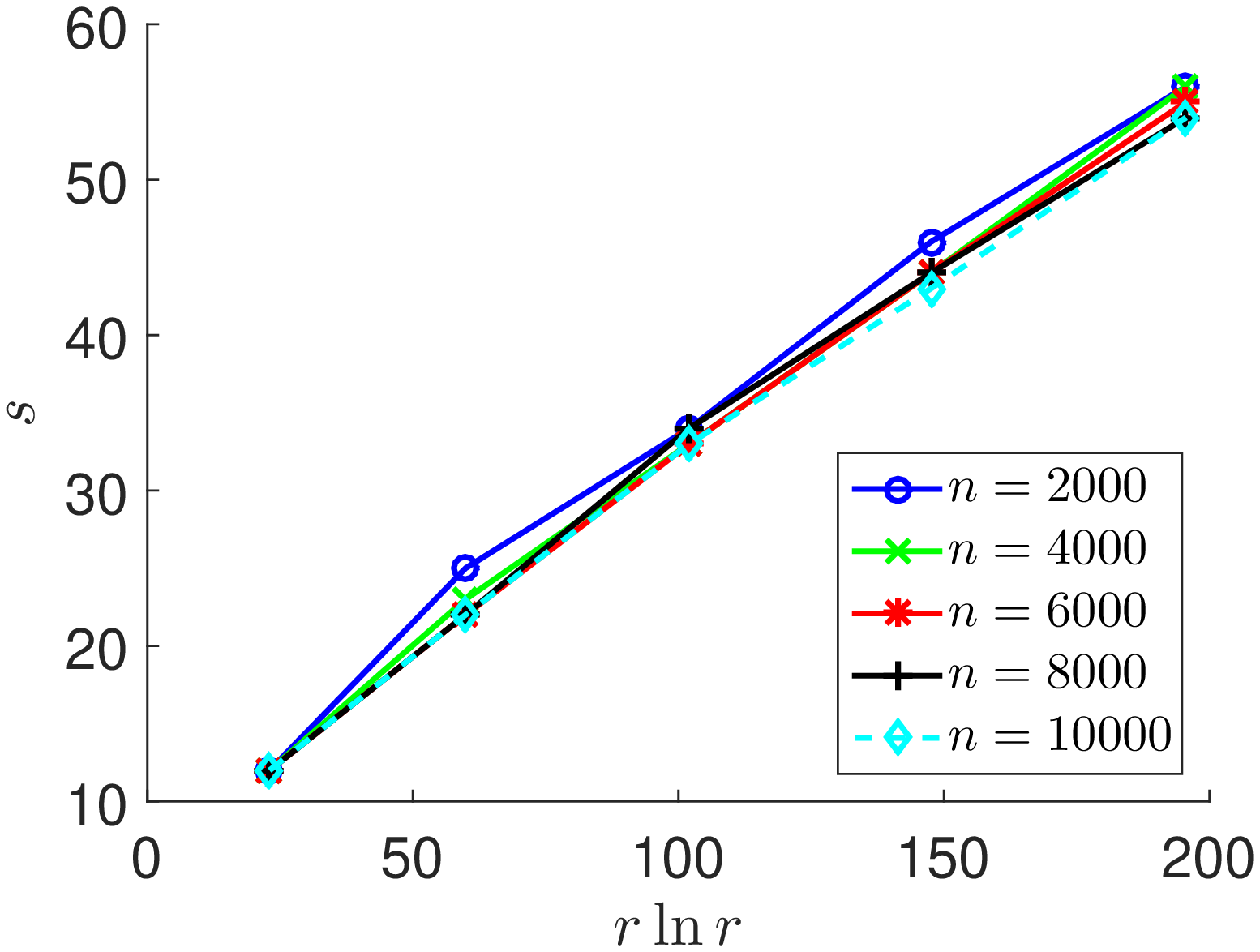}}
\centering
\subfigure[$s$ against $r^2\ln r$]{\includegraphics[width=0.45\textwidth]{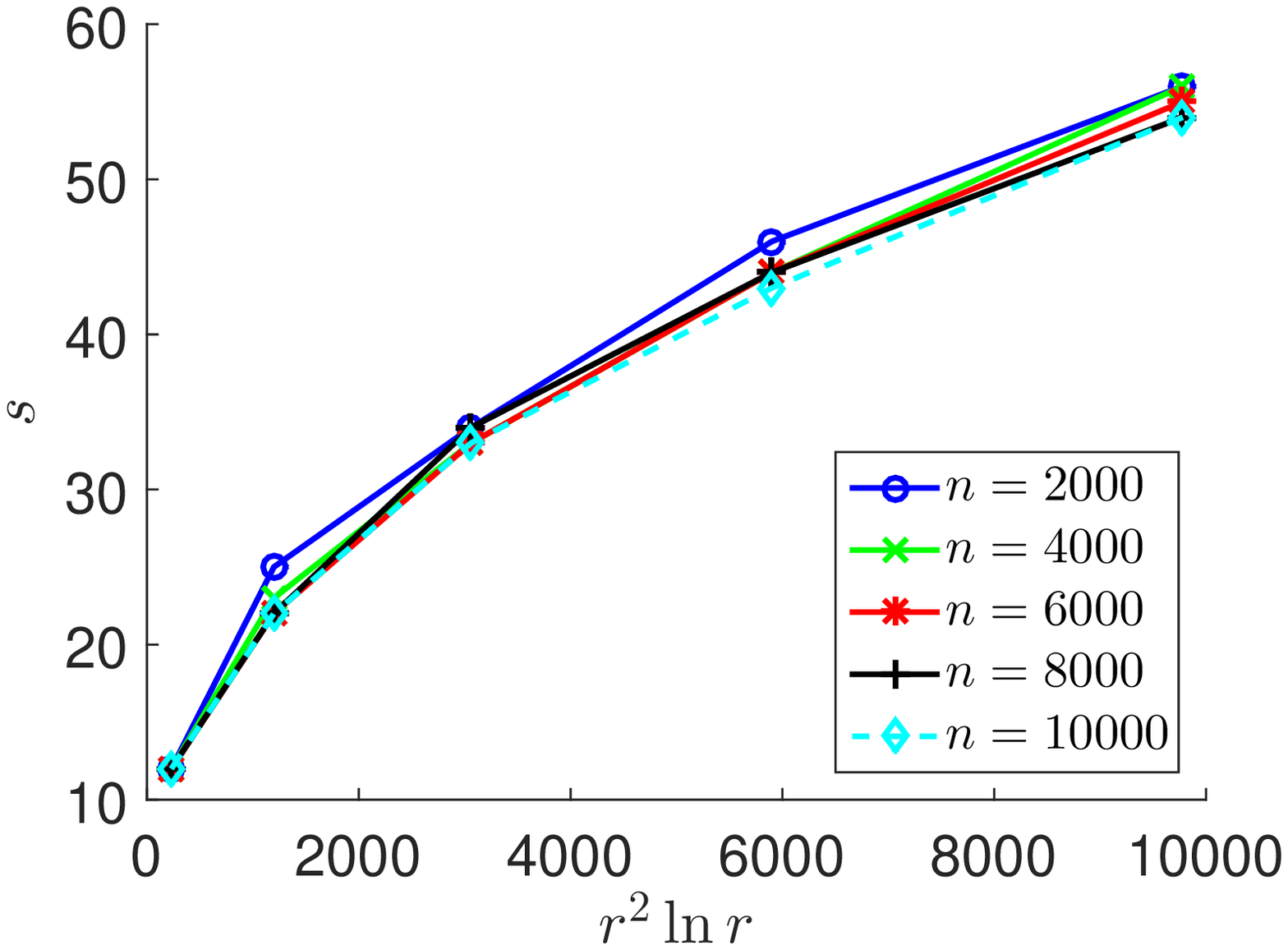}}
\caption{Experimental results of exact matrix completion on the synthetic data.} 
\label{fig1}
\end{figure}
\subsection{Comparison with Baseline Method for Low Rank Approximation}
Because Theorem \ref{thm2} shows that Algorithm \ref{LR} with uniform sampling enjoys additive error bound measured by Frobenius norm. To verify Theorem \ref{thm2}, we compare Algorithm \ref{LR} with uniform sampling against Nystr\"{o}m \citep{Relative:CUR} with uniformly sampled rows and columns to show that our algorithm with more general observation model can also perform well.  Since CUR$+$ requires the observed entries to meet more conditions, which limits its application, we do not include it in the comparison.
We note that algorithms based on approximate adaptive sampling and approximate norm sampling \citep{NIPS2013_Krishnamurthy,Wang2015} may yield better approximation, but these algorithms do not sample columns uniformly and can not deal with the unknown matrix with only one pass. Therefore, they are also not included in the comparison.
\begin{figure}[t]
\centering
\subfigure[$r=20,\sigma=0.1$]{\includegraphics[width=0.45\textwidth]{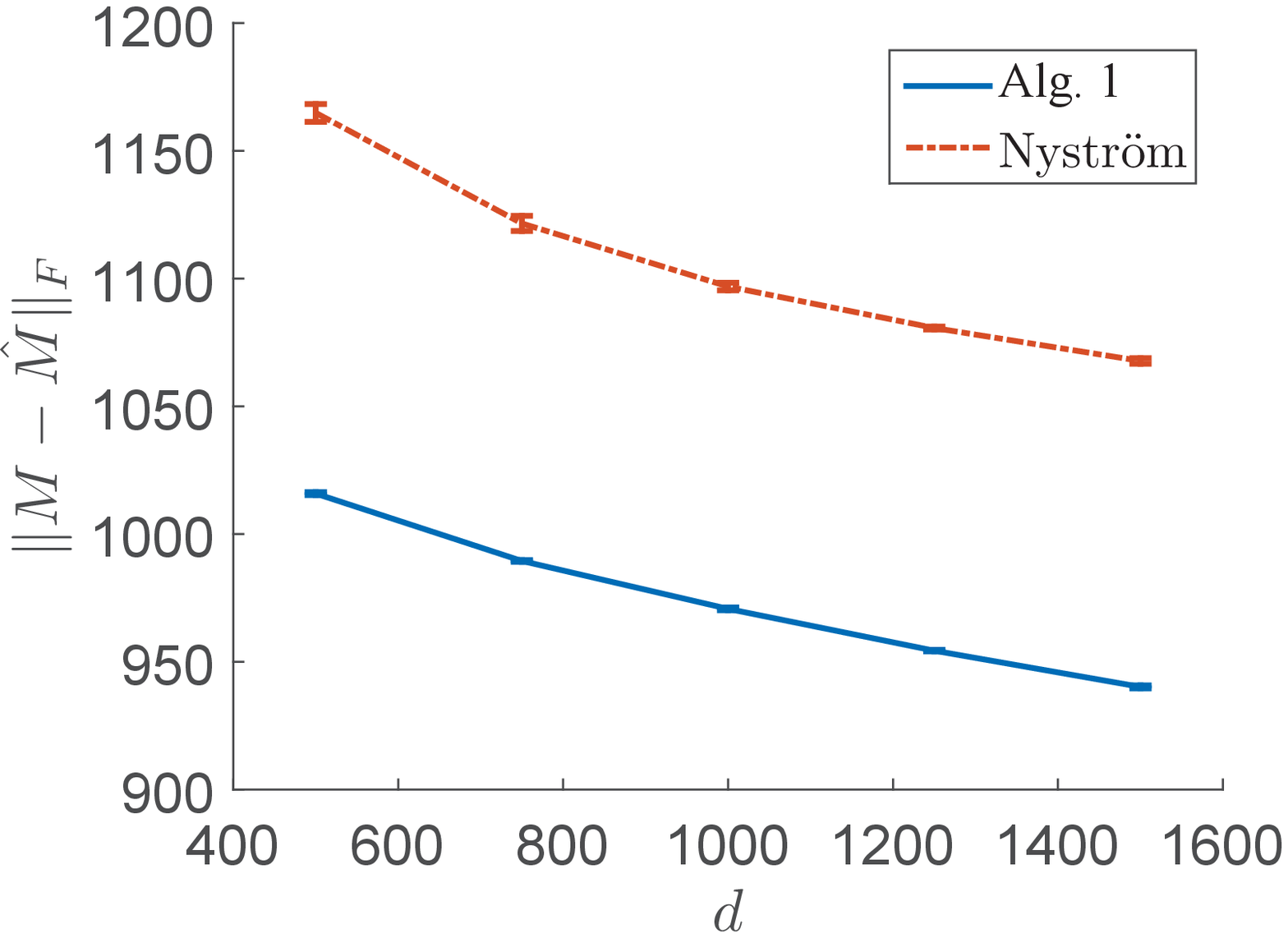}}
\centering
\subfigure[$r=20,\sigma=1$]{\includegraphics[width=0.45\textwidth]{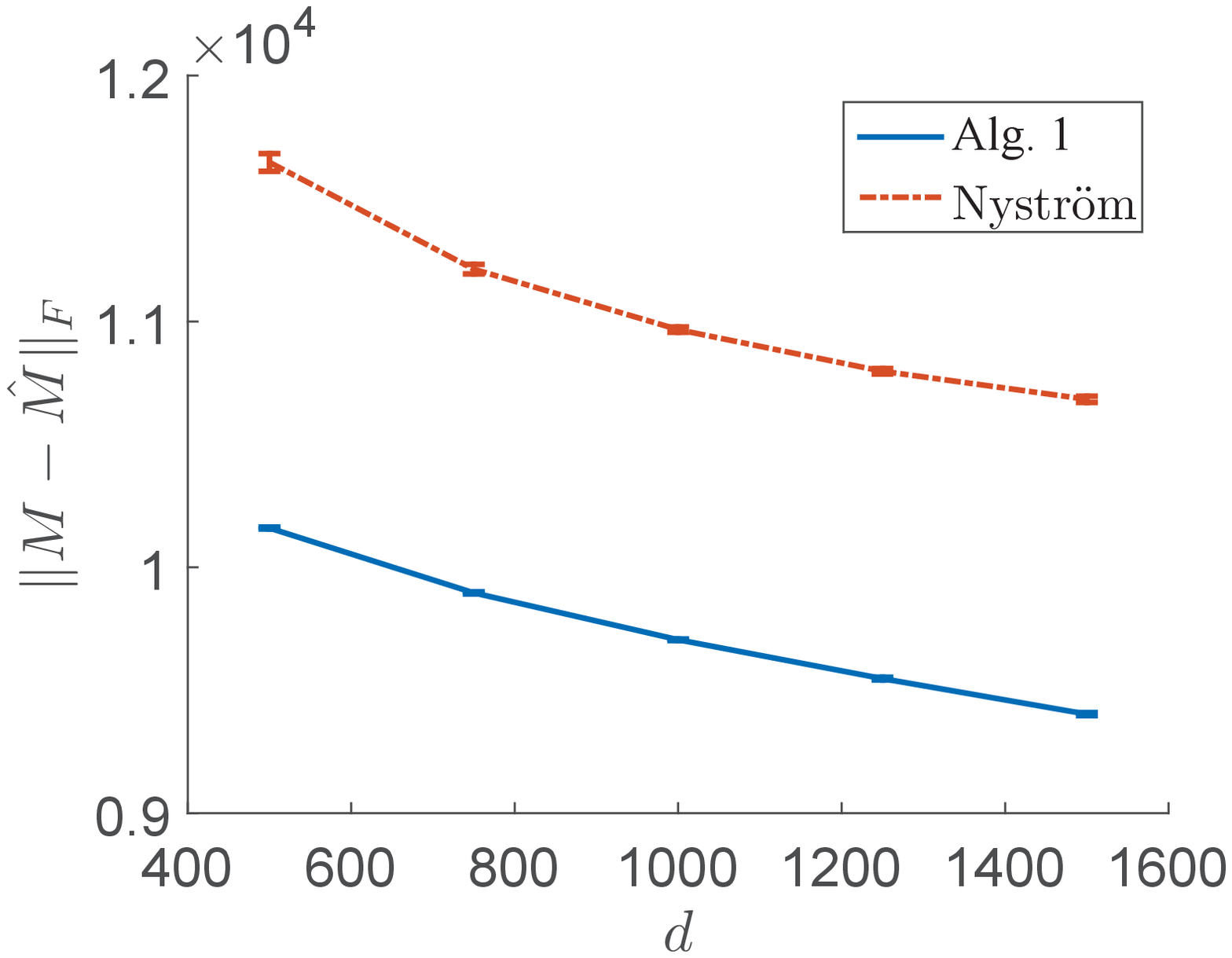}}\\
\centering
\subfigure[$r=40,\sigma=0.1$]{\includegraphics[width=0.45\textwidth]{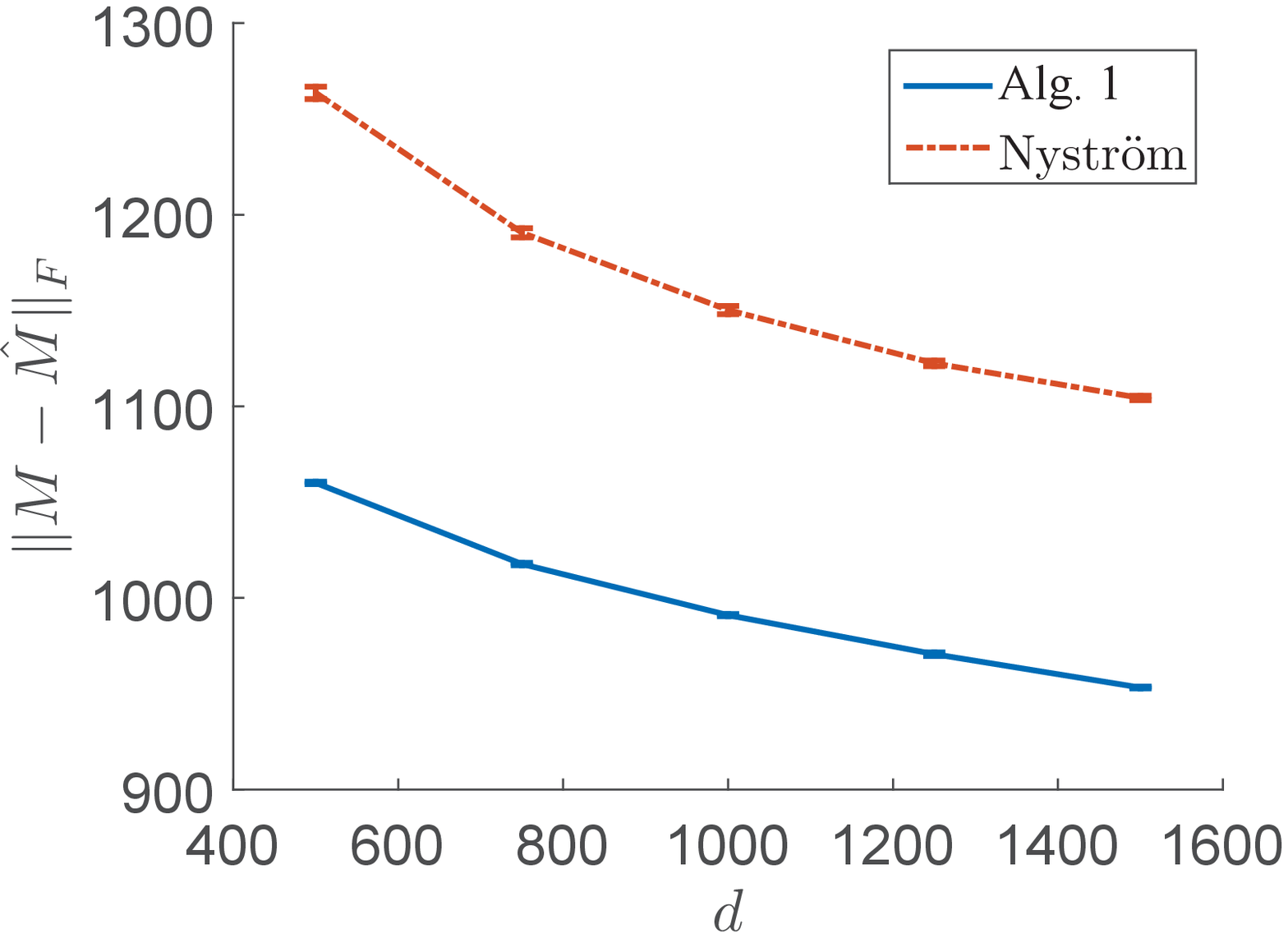}}
\centering
\subfigure[$r=40,\sigma=1$]{\includegraphics[width=0.45\textwidth]{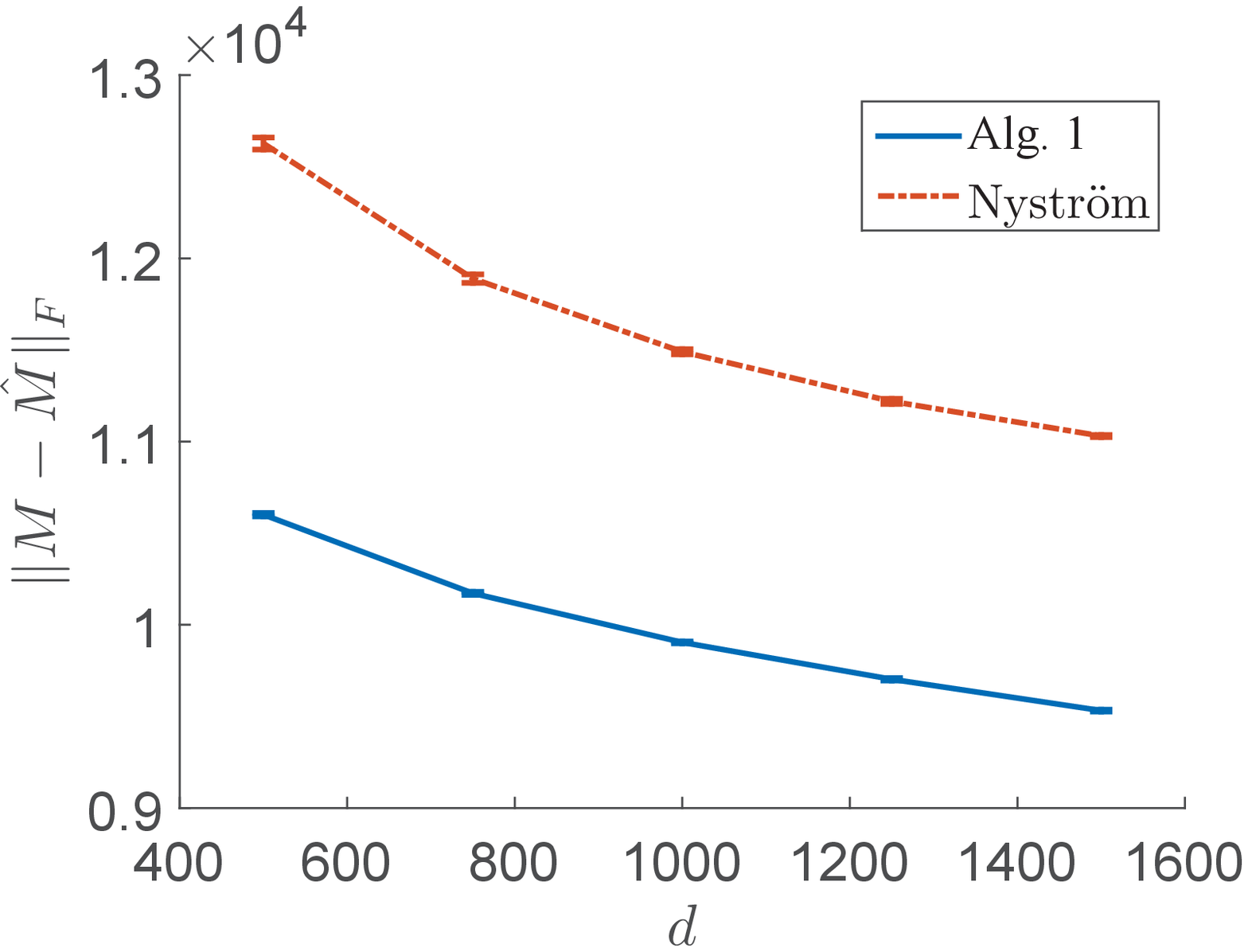}}
\caption{Experimental results of low rank approximation on the synthetic data.} 
\label{fig2}
\end{figure}
\paragraph{Settings} Following the experiment strategy of \citet{ICML2015_Xu}, we set the number of uniformly sampled columns and rows to $\alpha r$ and $\alpha r^2$ respectively for Nystr\"{o}m, where $r$ is the target rank and $\alpha$ is a parameter. For the sake of fairness, when our algorithm observed $d$ columns and $s$ entries for each partially observed column, we set $\alpha r+\alpha r^2=d+s$ such that Nystr\"{o}m observes the same number of entries with our algorithm.
According to Theorem \ref{thm2}, our algorithm should run with parameters $d=\Omega\left(r/\epsilon^2\right)$ and $s=\Omega(r^2/\epsilon)$. Because $r\ll n$ and $\epsilon$ is expected to be small such that $\Omega(r\epsilon)=\Omega(1)$, we set $s=d$ for our algorithm in the following experiments. We study square matrices of size $10000\times 10000$ which are the sum of a low-rank matrix $C$ and a Gaussian noise matrix $R$, where each entry of $R$ is drawn independently at random from $\mathcal{N}(0,\sigma^2)$. To create the rank-$r$ matrix $C\in\mathbb{R}^{10000\times 10000}$, we randomly
generate matrix $C_L\in\mathbb{R}^{10000\times r}$ and $C_R\in\mathbb{R}^{r\times 10000}$, where each entry of
$C_L$ and $C_R$ is drawn independently at random from $\mathcal{N}(0,1)$, and $C$ is given by
$C=C_L\times C_R$. Under this construction scheme, the incoherence $\mu(M)$ is small for different $r$ and $\sigma$ (from minimum 2.2929 to maximum 2.6792).

\paragraph{Results} We repeat each experiment $10$ times and report the average and the standard deviation of the approximation error $\|M-\hat{M}\|_F$ in Figure \ref{fig2}.
We find that our algorithm outperforms Nystr\"{o}m for different $r$ and $\sigma$, when they observe the same number of entries. Compared with Nystr\"{o}m, besides more general observation requirements, an additional advantage of our algorithm is explicitly extracting and exploiting the low-rank space of the sampled columns, which makes our algorithm more accurate and robust.

\section{Conclusion}
In this paper, we propose an algorithm to address the matrix completion problem from non-uniformly sampled entries. According to our theoretical analysis, our algorithm can perfectly recover a low-rank matrix with $\Omega(r/p_{\min} \ln n)$ observed entries, despite the probability distribution used to sample columns. Specifically, our algorithm with uniform sampling only needs $\Omega(r n \ln n)$ number of observed entries to perfectly recover a low-rank matrix. Furthermore, for noisy low-rank matrix, we show that the low-rank approximation computed by our algorithm with uniform sampling enjoys additive error bound measured by Frobenius norm. Numerical experiments verify our theoretical claims and demonstrate the effectiveness of our proposed algorithm.

\bibliography{ref}

\clearpage
\appendix
\section{Supplementary Analysis}
In this section, we first give the supporting theorems we will use in this analysis. Then we provided
the omitted proof.
\subsection{Supporting Theorems}
The following results are used throughout the analysis.


\begin{lem}
\label{Tropp-2}
(Theorem $1.1$ of \citet{Tropp2012}) Let $\mathcal{X}$ be a finite set of PSD matrices with dimension $k$ (means the size of the square matrix is $k\times k$). $\lambda_{\max}\left(\cdot\right)$ and $\lambda_{\min}\left(\cdot\right)$ calculate the maximum and minimum eigenvalues respectively. Suppose that
\begin{eqnarray*}
\max\limits_{X\in\mathcal{X}}\lambda_{\max}\left(X\right)\leq B.
\end{eqnarray*}
Sample $\{X_1,\cdots,X_l\}$ uniformly at random from $\mathcal{X}$ {independently}. Compute
\begin{eqnarray*}
\mu_{\max}=\lambda_{\max}\left(\sum\limits_{i=1}^l\E\left[X_i\right]\right), \mu_{\min}=\lambda_{\min}\left(\sum\limits_{i=1}^l\E\left[X_i\right]\right).
\end{eqnarray*}
Then
\begin{align*}
&\Pr\left\{\lambda_{\max}\left(\sum\limits_{i=1}^lX_i\right)\geq\left(1+\rho\right)\mu_{\max}\right\}\leq k\exp\frac{-\mu_{\max}}{B}\left[(1+\rho)\ln(1+\rho)-\rho\right] ~\text{for}~ \rho \geq 0,\\
&\Pr\left\{\lambda_{\min}\left(\sum\limits_{i=1}^lX_i\right)\leq\left(1-\rho\right)\mu_{\min}\right\}\leq k\exp\frac{-\mu_{\min}}{B}\left[(1-\rho)\ln(1-\rho)+\rho\right] ~\text{for}~ \rho \in[0,1).
\end{align*}
\end{lem}


\begin{lem}
(Lemma $1$ of \citet{Laurent2000})
\label{gau1}
Let $x\sim\chi_d^2$. Then with probability at least $1-2\delta$ the following holds
\[-2\sqrt{d\ln\left(1/\delta\right)}\leq x-d\leq2\sqrt{d\ln\left(1/\delta\right)}+2\ln\left(1/\delta\right).\]
\end{lem}

\begin{lem}
\label{gau2}
Let $x_1,\cdots,x_n\sim\mathcal{N}(0,\sigma^2)$. Then with probability  at least $1-\delta$ the following holds
\[\max_{i\in[n]}|x_i|\leq\sigma\sqrt{2\ln\left(2n/\delta\right)}.\]
\end{lem}

\begin{lem}
(Corollary $5.35$ of \citet{Vershynin})
\label{gau3}
Let $R$ be an $n\times t$ random matrix with independent and identically distributed standard Gaussian entries. Then for every $\epsilon\geq 0$ with probability at least $1-2\exp(-\epsilon^2/2)$ the following holds
\[\sqrt{n}-\sqrt{t}-\epsilon\leq\sigma_{\min}(R)\leq\sigma_{\max}(R)\leq\sqrt{n}+\sqrt{t}+\epsilon.\]
\end{lem}

\begin{lem}
\label{Drineas-2}
(Theorem $2$ in \citet{Drineas_2006_II})
Suppose $M\in\mathbb{R}^{m\times n}$. Let $A$ and $\hat{U}$ be constructed by Algorithm \ref{LR}. Then we have
\[\|M-\hat{U}\hat{U}^TM\|_F^2\leq\|M-M_r\|_F^2+2\sqrt{r}\|MM^T-AA^T\|_F.\]
\end{lem}

\begin{lem}
\label{smale}
(Lemma $2$ in \citet{smale2007})
Let $\mathcal{H}$ be a Hilbert space and let $\xi$ be a random variable with values in $\mathcal{H}$. Assume $\|\xi\|\leq M\leq \infty$ almost surely. Denote $\sigma^2(\xi)=\mathbb{E}\left[\|\xi\|^2\right]$. Let $\left\{\xi_i\right\}_{i=1}^d$ be $d \left(d<\infty\right)$ independent drawers of $\xi$. For any $0<\delta<1$, with confidence $1-\delta$
\[
\left\|\frac{1}{d}\sum_{i=1}^d[\xi_i-\mathbb{E}[\xi_i]]\right\| \leq \frac{2M\ln(2/\delta)}{d} + \sqrt{\frac{2\sigma^2(\xi)\ln(2/\delta)}{d}}.
\]
\end{lem}

\subsection{Proof of Lemma \ref{thm2-3}}
Let $i_1,\cdots,i_q$ are the $d$ selected columns. Define $S =
(\mathbf{e}_{i_1}/\sqrt{dp_{i_1}},\mathbf{e}_{i_2}/\sqrt{dp_{i_2}},\cdots,\mathbf{e}_{i_d}/\sqrt{dp_{i_d}})\in \mathbb{R}^{n\times d}$ where $\mathbf{e}_i$ is the $i$-th
canonical basis. Such that we have $A = MS$, that is, $A$ is composed of $d$ selected and rescaled columns
of $M$. Let the SVD of $M$ be $M=\bar{U}\bar{\Sigma} \bar{V}^T$, where $\bar{U}\in\R^{m\times r},\bar{\Sigma}\in\R^{r\times r},\bar{V}\in\R^{n\times r}$. We have $A=\bar{U}\bar{\Sigma} \bar{V}^TS$. To prove $\rk(A)=r$, we need to bound the minimum eigenvalue of $\Psi\Psi^T$, where
$\Psi=\bar{V}^TS\in\mathbb{R}^{r\times d}$. We have
\begin{eqnarray*}
\Psi\Psi^T=\bar{V}^TSS^T\bar{V}=\sum\limits_{j=1}^d\frac{1}{dp_{i_j}}{\bar{V}}_{(i_j)}^T{\bar{V}}_{(i_j)}
\end{eqnarray*}
where $\bar{V}_{(i)}$, $i\in[n]$ is the $i$-th row vector of $\bar{V}$.\\
It is straightforward to show that
\begin{eqnarray*}
\E\left[\bar{V}_{(i_j)}^T\bar{V}_{(i_j)}\right]=\sum_{i=1}^np_i\frac{1}{dp_{i}}\bar{V}_{(i)}^T\bar{V}_{(i)}=\frac{1}{d}I_r \text{ and } \E\left[\Psi\Psi^T\right]=I_r.
\end{eqnarray*}
To bound the minimum eigenvalue of $\Psi\Psi^T$, we need Lemma \ref{Tropp-2}, where we first need
to bound the maximum eigenvalue of $\frac{1}{dp_{i}}\bar{V}_{(i)}^T\bar{V}_{(i)}$, which is a rank-$1$ matrix, whose
eigenvalue
\begin{align*}
\max\limits_{i\in [n]} \lambda_{\max}\left(\frac{1}{dp_{i}}\bar{V}_{(i)}^T\bar{V}_{(i)}\right)\leq\frac{1}{dp_{\min}}\max\limits_{1\leq i\leq n}\|\bar{V}_{(i)}\|_2^2\leq\frac{1}{dp_{\min}}\mu(r)\frac{r}{n}
\end{align*}
and
\begin{align*}
\lambda_{\min}\left(\sum\limits_{j=1}^d\E\left[\bar{V}_{(i_j)}^T\bar{V}_{(i_j)}\right]\right)=\lambda_{\min}\left(\E[\Psi\Psi^T]\right)=1.
\end{align*}
Thus, we have
\begin{align*}
\Pr\left\{\lambda_{\min}\left(\Psi\Psi^T\right)\leq(1-\rho)\right\}\leq & r\exp\frac{-1}{r\mu(r)/(ndp_{\min})}[(1-\rho)\ln(1-\rho)+\rho]\\
=&r\exp\frac{-ndp_{\min}}{r\mu(r)}[(1-\rho)\ln(1-\rho)+\rho].
\end{align*}
By setting $\rho=1/2$, we have,
\begin{align*}
\Pr\left\{\lambda_{\min}\left(\Psi\Psi^T\right)\leq\frac{1}{2}\right\} \leq r\exp\frac{-ndp_{\min}}{7r\mu(r)}=re^{-ndp_{\min}/\left(7r\mu(r)\right)}.
\end{align*}
Let $d\geq7\mu(r)r(t+\ln r)/(np_{\min})$, we have $re^{-ndp_{\min}/(7r\mu(r))}\leq e^{-t}.$
Then, we have
\begin{align*}
\Pr\left\{\sigma_{\min}\left(\Psi\right)\geq\sqrt{\frac{1}{2}}\right\}=
\Pr\left\{\lambda_{\min}\left(\Psi\Psi^T\right)\geq\frac{1}{2}\right\} \geq 1 - e^{-t}.
\end{align*}
This means $\rk(\Psi)=r$, so $\rk(A)=\rk(\bar{U}\bar{\Sigma}\Psi)=r$.

\subsection{Proof of Lemma \ref{thm2-4}}
According the previous definition, $\hat{U}_{(j)}$, $j\in\O_i$ is the $j$-th row vector of $\hat{U}$.
We have
\begin{eqnarray*}
\hat{U}_{\O_i}^T\hat{U}_{\O_i} = \sum\limits_{j\in\O_i}\hat{U}_{(j)}^T\hat{U}_{(j)}.
\end{eqnarray*}
It is straightforward to show that
\begin{eqnarray*}
\E\left[\hat{U}_{(j)}^T\hat{U}_{(j)}\right]=\frac{1}{m}I_{\hat{r}} \text{ and } \E\left[\hat{U}_{\O_i}^T\hat{U}_{\O_i}\right]=\frac{s}{m}I_{\hat{r}}.
\end{eqnarray*}
To bound the minimum eigenvalue of $\hat{U}_{\O_i}^T\hat{U}_{\O_i}$, we need Lemma \ref{Tropp-2},
where we first need to bound the maximum eigenvalue of $\hat{U}_{(j)}^T\hat{U}_{(j)}$, which is a
rank-$1$ matrix, whose eigenvalue
\begin{align*}
\max\limits_{j\in [m]} \lambda_{\max}\left(\hat{U}_{(j)}^T\hat{U}_{(j)}\right)=\max\limits_{j\in[m]}\|\hat{U}_{(j)}\|_2^2\leq{\hat{\mu}\left({\hat{r}}\right)\frac{{\hat{r}}}{m}}
\end{align*}
and
\begin{align*}
\lambda_{\min}\left(\sum_{j\in\O_i}\E\left[\hat{U}_{(j)}^T\hat{U}_{(j)}\right]\right)=\lambda_{\min}\left(\E\left[\hat{U}_{\O_i}^T\hat{U}_{\O_i}\right]\right)=\frac{|\O_i|}{m}.
\end{align*}
Thus, we have
\begin{align*}
\Pr\left\{\lambda_{\min}\left(\hat{U}_{\O_i}^T\hat{U}_{\O_i}\right)\leq\left(1-\rho\right)\frac{|\O_i|}{m}\right\}\leq& {\hat{r}}\exp\frac{-|\O_i|/m}{{\hat{r}}\hat{\mu}({\hat{r}})/m}[(1-\rho)\ln(1-\rho)+\rho]\\
=&{\hat{r}}\exp\frac{-|\O_i|}{{\hat{r}}\hat{\mu}({\hat{r}})}[(1-\rho)\ln(1-\rho)+\rho].
\end{align*}
By setting $\rho=1/2$, we have
\begin{align*}
\Pr\left\{\lambda_{\min}\left(\hat{U}_{\O_i}^T\hat{U}_{\O_i}\right)\leq\frac{|\O_i|}{2m}\right\}&\leq
{\hat{r}}\exp\frac{-|\O_i|}{7{\hat{r}}\mu({\hat{r}})}=\hat{r}e^{-\left|\O_i\right|/7{\hat{r}}\mu({\hat{r}})}
\end{align*}
where with $|\O_i|\geq7\hat{\mu}({\hat{r}}){\hat{r}}(t+\ln {\hat{r}} +\ln n)$, we have ${\hat{r}}e^{-|\O_i|/7{\hat{r}}\mu({\hat{r}})}\leq
\frac{1}{n}e^{-t}$, that is
\begin{eqnarray*}
\Pr\left\{\lambda_{\min}\left(\hat{U}_{\O_i}^T\hat{U}_{\O_i}\right)\geq\frac{|\O_i|}{2m}\right\}\geq
1-\frac{1}{n}e^{-t}.
\end{eqnarray*}

\subsection{Proof of Lemma \ref{n1}}
According to our algorithm, we have selected $\hat{A}=[M^{(i_1)},M^{(i_2)},\cdots,M^{(i_d)}]$ from $M$ and rescaled it to $A$, which means $\spa(A)=\spa(\hat{A})$. Let $\hat{C}=[C^{(i_1)},C^{(i_2)},\cdots,C^{(i_d)}]$, $\hat{R}=[R^{(i_1)},R^{(i_2)},\cdots,R^{(i_d)}]$ and the SVD of $\hat{U}\hat{U}^T\hat{R}$ be $\hat{U}\hat{U}^T\hat{R}=U\Sigma V^T$, where $U\in\R^{m\times \hat{r}},\Sigma\in\R^{r\times \hat{r}},V\in\R^{d\times \hat{r}}$. Because of $\hat{A}=\hat{C}+\hat{R}$, we have \begin{align*}
\spa(\hat{U})&\subseteq\spa(\hat{U})\cap\spa(\hat{A})\\
&\subseteq\left(\spa(\hat{U})\cap\spa(\bar{U})\right)\cup\left(\spa(\hat{U})\cap\spa(\hat{R})\right)\\
&\subseteq\spa(\bar{U})\cup\spa(\hat{U}\hat{U}^T\hat{R}t)\\
&=\spa(\bar{U})\cup\spa(\hat{U}\hat{U}^T\hat{R}V)\\
&\subseteq\spa(\bar{U})\cup\spa(\hat{R}V).
\end{align*}
Let $\widetilde{R}=\hat{R}V\in\R^{m\times \hat{r}}$. Note that $\widetilde{R}$ is also a Gaussian random matrix because $V$ is orthogonal matrix and $\hat{R}$ is a Gaussian random matrix.
Consequently, with probability at least $1-2\exp(-\epsilon^2/2)-\delta/2$, we have the following bound on $\left\|P_{\hat{U}}\mathbf{e}_i\right\|_2^2$ as
\begin{align*}
\|P_{\hat{U}}\mathbf{e}_i\|_2^2\leq&\|P_{\bar{U}}\mathbf{e}_i\|_2^2+\|P_{\widetilde{R}}\mathbf{e}_i\|_2^2\leq\frac{r\mu(r)}{m}+\|\widetilde{R}\|_2^2\|(\widetilde{R}^T\widetilde{R})^{-1}\|_2^2\|\widetilde{R}^T\mathbf{e}_i\|_2^2\\
\leq&\frac{r\mu(r)}{m}+\frac{(\sqrt{m}+\sqrt{\hat{r}}+\epsilon)^2\sigma^2}{(\sqrt{m}-\sqrt{\hat{r}}-\epsilon)^4\sigma^4}\cdot\sigma^2\left(\hat{r}+2\sqrt{\hat{r}\ln(2/\delta)}+2\ln(2/\delta)\right)
\end{align*}
where $\|\hat{R}\|_2^2$ and $\|(\hat{R}^T\hat{R})^{-1}\|_2^2$ are bounded by Lemma \ref{gau3}, and $\|\hat{R}^T\mathbf{e}_i\|_2^2$ is bound by Lemma \ref{gau1}. The last inequality holds with probability at least $1-\delta$ by setting $\epsilon=\sqrt{2\ln(4/\delta)}$. Note that the fraction 
$(\sqrt{m}+\sqrt{\hat{r}}+\epsilon)^2/(\sqrt{m}-\sqrt{\hat{r}}-\epsilon)^4$  is approximately $O(1/m)$, when $\hat{r}\leq m/2$ and $\delta$ is not exponentially small (e.g., $\sqrt{2\ln(4/\delta)}\leq\frac{\sqrt{m}}{4}$). Then, with probability at least $1-m\delta$, we have
\begin{align*}
\hat{\mu}(\hat{r})=&\frac{m}{\hat{r}}\max_{i\in[m]}\|P_{\hat{U}}\mathbf{e}_i\|_2^2\leq\frac{r\mu(r)}{\hat{r}}+O\left(\frac{\hat{r}+\sqrt{\hat{r}\ln(1/\delta)}+\ln(1/\delta)}{\hat{r}}\right)\\
=&O\left(\frac{r\mu(r)+\hat{r}+\sqrt{\hat{r}\ln(1/\delta)}+\ln(1/\delta)}{\hat{r}}\right)\\
=&O\left(\frac{r\mu(r)\ln(1/\delta)}{\hat{r}}\right).
\end{align*}
Setting $\delta=\delta^\prime/m$, we prove (\ref{mu1}).

The projected vector $P_{\hat{U}^\bot}\mathbf{m}_i$ can be wrote as $P_{\hat{U}^\bot}\mathbf{m}_i=\hat{\mathbf{c}}+\hat{\mathbf{r}}$, where $\hat{\mathbf{c}}=P_{\hat{U}^\bot}\mathbf{c}$ and $\hat{\mathbf{r}}=P_{\hat{U}^\bot}\mathbf{r}$. By definition, $\hat{\mathbf{c}}$ lies in $\spa(\hat{U}^\bot)\cap\spa(\bar{U})$, and $\hat{\mathbf{r}}$ lies in $\spa(\hat{U}^\bot)\cap\spa(\bar{U}^\bot)$ with rank at least $m-r-\hat{r}$. Note that $\hat{\mathbf{r}}$ is still a Gaussian random vector. As a result, with probability  at least $1-\delta$, we have
\begin{align*}
\mu(P_{\hat{U}^\bot}\mathbf{m}_i)=& m\frac{\|\hat{\mathbf{c}}+\hat{\mathbf{r}}\|^2_{\infty}}{\|\hat{\mathbf{c}}+\hat{\mathbf{r}}\|_2^2}\leq 3m\frac{\|\hat{\mathbf{c}}\|^2_{\infty}+\|\hat{\mathbf{r}}\|^2_{\infty}}{\|\hat{\mathbf{c}}\|_2^2+\|\hat{\mathbf{r}}\|_2^2}\leq3m\frac{\|\hat{\mathbf{c}}\|^2_{\infty}}{\|\hat{\mathbf{c}}\|_2^2}+3m\frac{\|\hat{\mathbf{r}}\|^2_{\infty}}{\|\hat{\mathbf{r}}\|_2^2}\\
\leq&3r\mu(r)+\frac{6m\sigma^2\ln(4mn/\delta)}{\sigma^2(m-r-\hat{r})-2\sigma^2\sqrt{(m-r-\hat{r})\ln(2n/\delta)}}
\end{align*}
for all  partially observed $\mathbf{m}_i$, where $\|\hat{\mathbf{r}}\|^2_{\infty}$ is bounded by Lemma \ref{gau2} and $\|\hat{\mathbf{r}}\|_2^2$ is bounded by Lemma \ref{gau1}. Note that when $r\leq m/4$ and $\ln(2n/\delta)\leq m/64$, the denominator \[\sigma^2(m-r-\hat{r})-2\sigma^2\sqrt{(m-r-\hat{r})\ln(2n/\delta)}\geq\sigma^2m/4.\]
Subsequently, we have
\[\mu(P_{\hat{U}^\bot}\mathbf{m}_i)\leq3r\mu(r)+24\ln(2mn/\delta)\]
for for all partially observed $\mathbf{m}_i$.

\subsection{Proof of Lemma \ref{lemLR}}
Let $\xi_t = d\mathbf{a}_t\mathbf{a}_t^T$, where $\mathbf{a}_t$ is the $t$-th column of $A$ constructed by Algorithm \ref{LR} with uniform sampling. We have
\begin{align*}
\|\xi_t\|_F =\|d\mathbf{a}_t\mathbf{a}_t^T\|_F=n\|\mathbf{m}_{i_t}\|_2^2
\leq n\max_{i\in [n]}\|\mathbf{m}_i\|_2^2=\mu(M)\|M\|_F^2,\\
\E\left[\|\xi_t\|_F^2\right] = n\sum_{i=1}^n\|\mathbf{m}_{i}\|_2^4\leq\mu(M)\|M\|_F^4,\\
\E[\xi_t]=\sum_{i=1}^n\mathbf{m}_{i}\mathbf{m}_{i}^T=MM^T.
\end{align*}
According to Lemma \ref{smale}, with a probability $1-\delta$, we have,
\begin{align}
\label{MA}
&\|AA^T-MM^T\|_F=\left\|\frac{1}{d}\sum_{t=1}^d\left[\xi_t-\E[\xi_t]\right]\right\|_F \nonumber\\
\leq& \frac{2\ln(2/\delta)\mu(M)\|M\|_F^2}{d}+ \sqrt{\frac{2\ln(2/\delta)\mu(M)\|M\|_F^4}{d}}.
\end{align}
We complete the proof by substituting (\ref{MA}) into Lemma \ref{Drineas-2}.

\end{document}